\documentclass{article}
\usepackage{ijcai19}

\usepackage{times}
\usepackage{cite}
\usepackage{soul}
\usepackage{url}
\usepackage[hidelinks]{hyperref}
\usepackage[utf8]{inputenc}
\usepackage[small]{caption}
\usepackage{graphicx}
\usepackage{amsmath}
\usepackage{booktabs}
\usepackage{algorithm}
\usepackage{algorithmic}
\begin{document}
%\setlength{\belowcaptionskip}{-40pt}
\title{A general framework for scientifically inspired explanations in AI}

\author{
David Tuckey
\and
Alessandra Russo
\and
Krysia Broda
\affiliations
Department of Computing, Imperial College London\\
\emails
\{david.tuckey17, a.russo, k.broda\}@imperial.ac.uk
}


%\author{David Tuckey}
%\email{david.tuckey17@imperial.ac.uk}
%\affiliation{%
%  \institution{Imperial College London}
%  \country{United Kingdom}
%}

%\author{Alessandra Russo}
%\email{a.russo@imperial.ac.uk}
%\affiliation{%
%  \institution{Imperial College London}
%  \country{United Kingdom}
%}

%\author{Krysia Broda}
%\email{k.broda@imperial.ac.uk}
%\affiliation{%
%  \institution{Imperial College London}
%  \country{United Kingdom}
%}


\maketitle

\begin{abstract}
Explainability in AI is gaining attention in the computer science community in response to the increasing success of deep learning and the important need of justifying how such systems make predictions in life-critical applications. The focus of explainability in AI has predominantly been on trying to gain insights into how machine learning systems function by exploring relationships between input data and predicted outcomes or by extracting simpler interpretable models. Through literature surveys of philosophy and social science, authors have highlighted the sharp difference between these generated explanations and human-made explanations and claimed that current explanations in AI do not take into account the complexity of human interaction to allow for effective information passing to not-expert users. In this paper we instantiate the concept of structure of scientific explanation as the theoretical underpinning for a general framework in which explanations for AI systems can be implemented. This framework aims to provide the tools to build a "mental-model" of any AI system so that the interaction with the user can provide information on demand and be closer to the nature of human-made explanations. We illustrate how we can utilize this framework through two very different examples: an artificial neural network and a Prolog solver and we provide a possible implementation for both examples.
\end{abstract}

\section{Introduction}
\label{sec:introduction}

The recent surge in work on explainable artificial intelligence (xAI) responds to an increasing need for explaining complex machine learning systems. These systems now aim to tackle complex tasks usually accomplished by humans, raising legal and ethical concerns. A line of research from the machine learning (ML) community aims to obtain an insight into the "black-box" through a multitude of techniques\cite{Guidotti2018} generating either an interpretation of the behaviour of the system or a simpler understandable representation. This approach focused on the ML system is now criticized\cite{Rudin2018, Ribera2019} due to the lack of attention given to the final user and suggestions are made to include considerations from social science in xAI\cite{Miller2019, Mittelstadt2018}. Indeed human explanations are very complex, based on the cognitive biases of both the explainee and explainer. These biases should be taken into account when generating explanations.

One difficulty in xAI is that no consensus has been attained on the definitions of "explanation", "interpretation", "explainability" and "interpretability" in spite of work to clarify their use\cite{Lipton2016, Doshi2017, Gilpin2018}. We will call, in this paper, \textit{explanation process} the action of explaining while explanations, as a product\cite{Miller2019}, can either be individual or composed. An individual explanation is an answer to a single question while a composed explanation is the result of a dialogue between explainer and explainee and is composed of multiple individual explanations (see Figure \ref{fig:example_explanation}). We will call the AI inference system \textit{AI system} and reserve the use of the word \textit{model} to the meaning defined in Section \ref{sec:overton} to avoid confusion.

\begin{figure}[h]
    \framebox{\parbox{\dimexpr\linewidth-2\fboxsep-2\fboxrule}{
    S : "Why did the ball fall down ?"\\
    T : "You dropped it so it was in a free fall attracted by the earth."\\
    S : "Why did the earth attract it ?"\\
    T : "The laws of gravity state that massive objects like the earth attracts object around it, by the force called gravity."
    }}
    \caption{Example of composed explanation between a teacher and a student, composed of multiple individual explanations.}
    \label{fig:example_explanation}
\end{figure}

Instead of looking into how to explain an AI system's prediction, we focus on how to explain an AI system's prediction \textit{to a user}. We aim to create a system capable of giving composed explanations to a user who doesn't have any previous knowledge of the system's functioning, essentially recreating the same interaction type as shown in Figure \ref{fig:example_explanation} for any AI system. Miller\cite{Miller2019} argues that explanations are social and selective, meaning that they are the result of an interaction and biased to only present a chosen amount of information.  This requires to take into account the different types and level of explanations\cite{Miller2019}. For example, when considering an object falling, we understand that there is a difference between explaining that the operator dropped it and explaining that it fell because it was subjected to the laws of gravity. The first explanation takes a causal approach while the second one abstracts the event to give a more general justification. Generating such explanations requires the explainer to have a good understanding of the phenomena at hand, both on a concrete and abstract level. We will call this understanding the "mental-model" that the explainer has and uses to reply to the explainee's questions.

We propose to use the structure of scientific explanation as defined by Overton\cite{Overton2012} to organize and implement in an explanation system the mental-model of a given AI system we wish to explain to the user. The explainer (the explanation system) then interacts with the explainee (the user) using knowledge from the mental-model to answer the questions in a dialogue. We provide two implemented examples of mental-models for AI systems. The choice of the structure of scientific explanations allows for an agnostic representation, meaning that we can represent very different AI systems using the same framework. We also assumed that the user would be more sensitive to scientific explanation than to AI-specific explanations. This work follows the suggestion made by Miller\cite{Miller2019} to take inspiration from the social sciences.

This paper is composed as follows: Section \ref{sec:related} cites a few related work, Section \ref{sec:overton} presents the structure of scientific explanation as defined by Overton\cite{Overton2012} and we explain in Section \ref{sec:adaptation} how we adapted it in the context of AI. We then present our implementation of the idea into a framework that could be used for any AI system in Section \ref{sec:framework}. We discuss our work in Section \ref{sec:discussion} before concluding in Section \ref{sec:conclusion}.

\section{Related Work}
\label{sec:related}
Explanations were already an important part of information systems before recent deep learning advances. Some work used a similar approach consisting of choosing a theoretical structure as the basis of a framework for human-machine interaction. Johnson and Johnson\cite{Johnson1993} derived from the Task Knowledge Structure (TKS), a theoretical framework to task analysis, models of the knowledge required to provide explanations. For a given task, one can derive the TKS of it and then use this model to construct explanations. 

Yetim\cite{Yetim2008} proposes a framework to organize justification by splitting explanation types using Habermas\cite{Habermas1984} discourse theory, which splits arguments into different types: explicative, theoretical, pragmatic, ethical, moral, legal, aesthetic and therapeutic. Yetim then uses Toulmin's\cite{Toulmin2003} argumentation schema to create a concrete structure for each discourse type. Using this framework to structure the construction of explanation, a system can provide the user with an appropriately structured justification depending on the context.

%\section{Related Works}
\section{The Structure of scientific explanations}
\label{sec:overton}

\begin{figure*}[t]
    \centering
    \includegraphics[width=\textwidth]{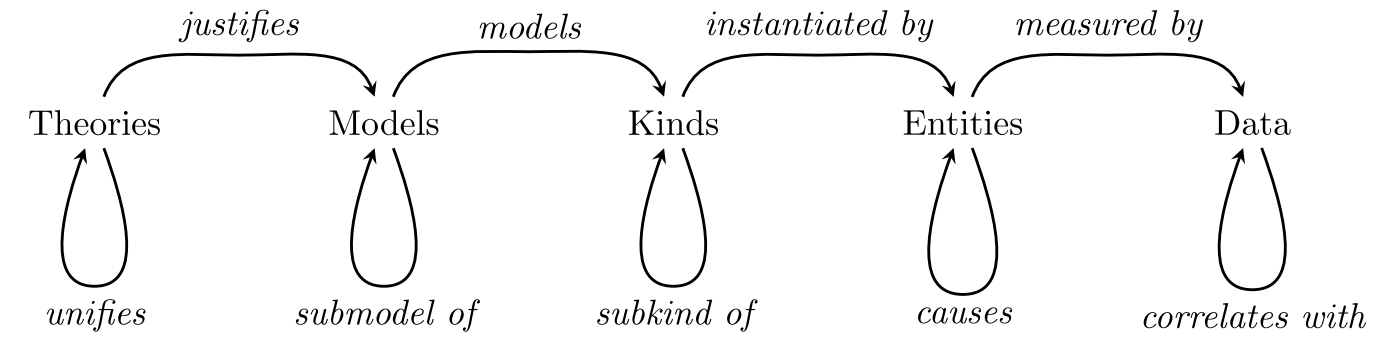}
    \caption{The five categories of explanan and explanandum in the structure derived by Overton, together with the main relations between categories. Reproduced from \cite{Overton2012}.}
    \label{fig:overton_structure}
\end{figure*}

We present in this section the structure of scientific explanation as defined by Overton\cite{Overton2012}. He defines an explanation being composed of an explanan, an explanandum and of the relation that links the two: the explain-relation. Explanan and explanandum can be of five categories : \textit{data, entities, kinds, models} and \textit{theories}. The explain-relation is then a link between two elements of these categories. A visual representation is given in Figure \ref{fig:overton_structure}. To fluidify the presentation of this structure, we will be illustrating the different categories with a simple example, also highlighting the generality of the structure. Our example will be one of a dog, Ralph, burying a bone in a garden.

\subsection{The five categories of explanan/explanandum}

\paragraph{Data:}

Data are concrete statements about entities. They are the result of measurements or observations. It is important to note that data are about concrete objects (entity). 

In our example, the data might correspond to the video recording of the dog burying the bone, the GPS coordinates of the location or the temperature/humidity reading in the garden.

\paragraph{Entities:}

An entity is a particular existing inside of a space-time dimension that has causal power in the physical world. It is created, undergo changes and disappears at some point. They are the objects of our observations and measurements: we create statements about them. It is easy to understand that physical objects (living organisms, inanimate objects...) are entities. Also are comprised in this category more abstract objects, such as specific processes that are observable, time periods (e.g the Jurassic), regions of space (e.g  a specific polluted patch of the ocean) or collections of entities(e.g a solution of sodium). 

Our dog, Ralph, is one entity together with the specific bone being buried and garden in which the scene is taking place. We do not have a specific name to designate the bone and garden and usually, in English, we only use the definite article with the name of the group to designate entities (e.g "the garden" in contrast to "a garden"). 

\paragraph{Kinds:}

Kinds are any abstract universals with which we classify entities. We could say it is the nature of the particular object considered. In our language, we use such things regularly ("car", "dog"). Even if we understand that two cars are not identical, we recognize that they share similarities which we mean by the appellation "car". A kind thus has characteristics, that Overton calls qualities or physicists call properties. Some kinds are easier to grasp than others: we understand well the concept of a dog. However, a person that is not eligible for credit is part of a kind that we could define as being "person not eligible for credit". Entities can then be seen as \textit{instances of kinds}. 

In our example, Ralph is of the kind "labrador" which is a subkind of "dog". The bone is an instance of the kind "bone" and the garden an instance of the kind "garden".

\paragraph{Models:}

"Model" takes here the meaning it has in physics: it is a set of rules that explain complex relationships between qualities of one or multiple kinds. The model is a "model of a kind". It describes the behaviour of how instances of kinds change when interacting with one another. A model can include a set of mathematical equations, a graph, a natural language description of a phenomenon...

The models in our example could be the behavioural model of dogs which describes generally how dogs behave (and bury bones) or models describing how the bone decomposes when buried. 

\paragraph{Theories:}

Overton defines theories as being an abstraction of the models, though he states that the line between the two is often blurry. Models are made from theories. Models are models of a kind, and theories are more general concepts stating more general rules about the world. The Newtonian laws can be seen as a theory from which we derive models (equations) of how kinds interact. 

In our example, theories are harder to define. One could argue that you can explain the behaviour of dogs by looking into the theory of evolution, making the latter a member of the theories category. 

\subsection{Explain-relation}

The last element of an explanation is the explain-relation. It specifies what the relation is between the explanan and the explanandum. Its nature can be defined by the nature of the explanan and explanandum: it can be an Entity-Entity relation, a Kind-Data relation, a Model-Entity relation, a Data-Theory relation... Overton calls this pair of categories the \textit{form} of the explanation for a total of 25 different forms. He defines four of these relations as being primary: Theory-Model, Model-Kind, Kind-Entity, and Entity-Data which correspond to justification, modelling, instantiation, and observation respectively. The rest of the relations are compositions of these four. We will present a few relations : 

\paragraph{Entity-Data:}
The link between entities and data is observation or measurement. This relation links an entity to a statement about it.

For example, a relation exists between Ralph burying its bone and the video recording of it, which is the action of filming the scene: the observation.

\paragraph{Entity-Entity:}
There are a large number of causal relationships that can hold between entities. Overton highlights parthood and singular causation. The former is the inclusion relation on a physical sense (e.g. a leg on a body), the later is a simple causal event of any form.

Ralph caused the bone to be buried underground, there can be a link between the bone's location and the dog.

\paragraph{Kind-Entity:}
The relation between Kinds and Entities is instantiation. An entity is an instance of a kind. Instantiation can lead to different entities, and this instantiation is often implied and not too relevant from an explanation point of view. It is however important to make the difference between kinds and entities. When one observes a chemical reaction, they use the name of the kind (e.g. $CO_2$) to designate the entities (i.e. the actual molecules in the laboratory) they are manipulating. 

Ralph is an instance of the kind "dog", the bone of the kind "bone", the garden of the kind "garden", so there exists an instantiation relation between these kinds and their respective instances.

\paragraph{Model-Kind explanation:}
Model-Kind relations consist of using a model to explain the changes in a kind. Overton noticed that these explanations are one of the most prevalent in science probably because of the prediction power of models. 

In our example, stating that dogs bury their food in a stressful environment to keep for later is a Model-Kind explanation.

\paragraph{Model-Entity explanation:}
This is a composition of a Model-Kind and Kind-Entity relation. It is the use of a model to explain the behaviour of a particular entity. 

Dogs burying their food in stressful environment explains why Ralph buried his.

\paragraph{Theory-Model explanation:}
Theories are used to build and justify models. Models have to refer to a more general mechanism that is shared between models.

\paragraph{Example}

By the way of example, let's consider the case of students adding sulfuric acid ($H_2SO_4$) to water (dilution) and observing on a pH-meter the change in pH. The numerical value displayed on the machine is the data. The entities involved here are the specific ions and molecules present in the different beakers used in the experiment. The kinds are the different types of molecules and ion involved ($H_2O$, $H^+$, $SO_4^{2-}$, $H_2SO_4$). The models are pretty simple here: a simple link between the concentration of ions $H^+$ and the pH and how $H_2SO_4$ splits into $H^+$ and $SO_4^{2-}$. Behind all these are the theories of the composition of matter as atoms and molecules together with chemical reactions. 

The question "Why is the pH going from 7 to 5 ?" asks a question about the measurement of pH of the solution, which is data. "Because of the presence of more ions $H^+$ in the solution" is an Entity-Data explanation while "Because $pH = -log[H^+]$" is a Model-Data explanation.

\section{Adaptation to Artificial Intelligence}
\label{sec:adaptation}

We present in this section how we use the structure of scientific explanation to generate explanations for AI systems. We will use the example of a neural network classifying images of animals to simplify our argumentation, but the same considerations hold for any other system. We begin by explaining the heuristic behind our approach to defining AI systems' mental-model using the structure of scientific explanation and then describe our interpretation of it.

\subsection{Considering AI systems as entities}

The naive interpretation of the structure for our classifier would be to consider the classification classes as the kinds, the individual animals photographed as the entities and the pictures as data. The models and theories would be more complicated to define especially if we want to include only concepts related to an image recognition task.

Let's assume that the system classified an image as being one of a cat. The question by the human operator would probably look like \textbf{"Why is this a cat ?"}. We here need to be careful of what is the implied explanandum: we could define it as being \textbf{"This is an image of a cat"}. This is the natural assumption that is made when two humans interact together, but we argue it is incomplete in the context of AI. Indeed, in AI the proper explanandum would be \textbf{"The AI system classified the image as being one of a cat"}. This comes from the fact that we can't assume in AI the explainer and explainee share similar knowledge so we need to question the inference process as well the nature of the input. We focused in our work on how to represent the inference process inside the structure of scientific explanation. 

Scientific explanation being mainly meant to explain observed phenomena, we propose to consider the prediction made by the AI system as being a "physical phenomenon" we observe and wish to explain. This analogy aims for an easier description of the categories for explanan and explanandums. Our approach suggests breaking the AI system's algorithm into entities that interact with each other, being instances of kinds ruled by models.  

\subsection{Definition of the five categories}

We are considering a prediction of an AI system and are trying to build an explanation for it. We observe (measure) the AI system during inference and consider it in the same way we would consider a ball falling. Our example of a neural network is a straightforward one due to the structure of the algorithm itself.

\paragraph{Data:} The data category is composed of the input of the AI system, its output, and all the intermediate hidden variables that have been computed by the system. 

For our neural network, the data will be the record of the activation of all of the neurons, the input image and output predicted class. We also count in the values of the parameters. To be fully complete, one could add the training data. 

\paragraph{Entities:} The specific AI system running the prediction is the main entity to consider. It's constituent (parameters, neurons, clauses, predicates, branches...) are also entities. We consider the system at the moment of prediction here, meaning at a specific parametric state. Even if it is not an entity in the physical sense, it is a particular that is in this state for a certain period of time and will be deleted at some point or transformed when retrained or modified.

For our neural network, we will have the input image, all the neurons, the parameters and the output of the system. When considering two networks that have the exact same architecture but are in different parametric states, they are two distinct entities that are instances of the same kind which describes their architecture. 

\paragraph{Kinds:} The kind category is composed of the types of AI systems, architecture-wise, that are instantiated into kinds. In these kinds, we can cite the artificial neural networks architectures, linear regression, Bayesian networks or SVM. Are also in the kind category the elements of these architectures, such as the neurons and the parameters for a neural network or the probability matrices and nodes for a Bayesian network. 

There are multiple ways of classifying entities for a neural network. One could have a kind for the architecture of the network, one for neurons and one for parameters. We could go further and define a kind for each specific neuron in the architecture (e.g. first neuron of the first layer), each neuron in a network would then be an instance of a specific kind describing their position. Such a specific definition can be useful for models and implementation purposes. 

\paragraph{Models:} We remind the reader that a model is a story of how a kind and its properties behave. We include the algorithm of the AI system broken down into individual interactions between kinds. The training algorithms are thus also comprised in the model categories. They describe (mathematically or not) the story of how the characteristics of the kinds are modified during training/inference. For example, a model of the training for a neural network is the partial derivative of all of the parameters in regards to the loss, or for a SVM the optimization task. 

For our example, we break down the forward pass of a neural network layer by layer or neuron by neuron. We define a model for computing the activation of each neuron (an equation of the form $g(Ax+b)$) from the parameters and the previous layer neurons' activation. The partial derivative of the loss function in regard to each parameter is part of the training models for the parameters, which could be included. 

\paragraph{Theories:} This category is not so complicated to define in the context of AI. Concepts like optimization or gradient descent are the building blocks of many \textit{models} of \textit{kind} of AI system. 

Back-propagation and gradient descent can be seen as a theory shared by all deep gradient-based AI systems and the idea used to derive the specific algorithms and equations for a neural network.

\subsection{Explain-relation}

Explanations are built from links between elements of these categories. We find in majority two types of explanations: Entity-Entity explanations and Model-Entity explanations. Indeed we aim to give two main types of explanation for AI: how a particular entity came to this instance by highlighting the characteristics of other entities, so a more "causal" approach; or present the part of the algorithm to the explainee that was responsible to compute the subject of the question. 

For our neural network example, we explain the activation of a specific neuron by highlighting the activation of the neurons in the previous layer with neuron to neuron relations (between entities) or provide the algorithm (model) that was used to compute the activation value with Model-Entity relations. The reader might discuss here the usefulness of such explanations as they are not helping the user to trust the result. We aim by this to allow the user to get information on the system's function and are not directly helping interpretation. We discuss these concerns in Section \ref{sec:discussion}.

\subsection{Examples}
We illustrate here further the content of these categories. We however would like to point out that there is not one way to define each example in this structure, depending on how complex we make the mental-model of the AI system. We will give more details as to our implementation in Section \ref{sec:framework_examples}.

\subsubsection{Neural network with interpretation}
% Model is the model of how the AI got to this representation from the training data. How the computations are done inside the network. The story can be very high level, like "why do you think that cats have 4 legs", "because it was in 97% of the data"
% Theory is what the model is based from, like backprop, or stuff like that
We can make the neural network example more complex by adding an interpretation technique like Layer-Wise Relevance Propagation\cite{Bach2015} (LRP). To the data we add the saliency map generated together with the relevance value of all the neurons. To the entities we add the saliency map and a numerical field to each neuron to represent the relevance. To the kinds we add the concept of saliency map and update the concept of neurons. The models now include the specific equations that were used to compute LRP through the network and the theories behind this technique. 
We also add explain-relations : from the saliency map to the output of the model representing the heatmap interpretation of what was important in the input, from neuron to neuron to signify that one neuron's activation/relevance participated in the other neuron's relevance and from parameter to neuron showing that the parameter's value participated in the computation of the neuron's relevance. 

\subsubsection{Prolog}

We now take the example of a symbolic system like Prolog\cite{Colmerauer1990}. The data is the result of the query and the trace. The entities will comprise all the terms, predicates, rules and variables in the program that is executed and queried. In the kinds are the concepts of terms, predicates, rules and variables. In the models are the different part of the SLDNF derivation implemented in the solver. The theory category includes top-down derivation, logic...

The types of relations will depend here on how complex and truthful they are to the real AI system. One can define simple relations like a predicate being evaluated to true because it unified with the head of rule, and all the predicates in the body of this rule evaluated to true. This is an Entity-Entity relation between a rule and a predicate, which does not represent exactly how the SLDNF algorithm functions but is a valid way of explaining the process. 

\section{Implementation as a general framework}
\label{sec:framework}

We present in this section our use of the structure of explanation as the foundation for a general framework to implement explanations. Our implementation tries to follow the pattern offered by the structure of scientific explanation, in particular the fact that Entities are instances of Kinds. We will call "explanation system" the software implementing the mental-model of the AI system and interacting with the user. We first present the general workflow that the framework requires and go into more details afterwards on how we approached the different problems.

\subsection{Workflow}

The first step to using our framework is to implement the mental-model of the AI system that we want the explanation system to query. We need to implement the kinds of each of the elements involved (see Section \ref{sec:framework_kinds}), the Entity-Entity relations (see Section \ref{sec:framework_links}) and the models (see Section \ref{sec:framework_models}). At runtime, we instantiate the kinds to generate the entities, all the Entity-Entity relations and set the values of the different characteristics of the entities to match the values computed by the AI system making the prediction. The user then interacts only with the explanation system if he requires any explanation on the output. We present the final interaction pattern in Section \ref{sec:framework_interaction} and explain how our explanation system runs a search for explanation in Section \ref{sec:framework_search}.

\subsection{Defining the kinds}
\label{sec:framework_kinds}

The most appropriate way we found to implement Kinds is as Classes in Oriented Object Programming to instantiate them in the running program to represent the Entities. We can then define Models on the Kinds and relate them easily (implementation-wise) to the Entities. The classes need to include all of the information needed about a Kind and its instantiation as an Entity. We define a Kind as being a set $(N, C, V)$ : 

\begin{itemize}
    \item $N$ : The name of the kind which is the name of the Class.
    \item $C$ : A set of constants that defines characteristics that have the same value for all of the instances of the kind.
    \item $V$ : A set of placeholders that defines the characteristics that are common to all instances of a kind but can take different values. We also add a name placeholder to give an identifying name to each entity.
\end{itemize}

\subsection{Defining the Entity-Entity relations}
\label{sec:framework_links}

We require to define by hand all the Entity-Entity relations. They are supposed to represent causal relations between parts of the AI system and provide a simple explanation as to what the relation is. They can be seen as putting an explanan entity in relation with an explanandum entity with a given reason as to how the explanan impacted the explanandum. It is important to note that they link entities and not kinds, so each relation needs to be instantiated at runtime with the entities. Thus we implemented these relations as classes that get instantiated. We define Entity-Entity relations as being a set $(N, En_t, Ed_t, En, Ed, R, P)$ :

\begin{itemize}
    \item $N$: The \textit{name} of the link, which is the name of the Class
    \item $En_t$: The \textit{type of the explanan}, which provides the kind and characteristics of the explanan that play a role in this relation
    \item $Ed_t$: The \textit{type of the explanandum}, which provides the kind and characteristics of the explanandum that play a role in this relation
    \item $En$: The \textit{explanan}, a pointer to the explanan entity
    \item $Ed$: The \textit{explanandum}, a pointer to the explanandum entity
    \item $R$: The \textit{reason}, a natural language information as to the nature of the relation
    \item $P$: The \textit{priority} is a bias that we hard-code in Entity-Entity relations allowing to set a preference between two different relations between the same entities, the one with the higher priority being returned first. 
\end{itemize}

\subsection{Defining the models}
\label{sec:framework_models}

One difficulty of defining such a framework is to present the user with an intelligible representation of a model. The other pre-requisite is to represent models in such a way that the program can automatically choose which one is appropriate for an explanation. Models can take many different forms: they can be natural language, equations or sets of equations, graphs, etc; the common part being that they describe how characteristics of a kind change when in contact with other kinds: it is a "story" of how things should happen. Models then have a context, which comprises all the kinds that are involved in the model, a result which is what specific kind changed as a result of the interaction and a "story" which is the modular part which can be of any form. To do so, we defined models as being a set $(N, C, R, MoF, S)$ : 

\begin{itemize}
    \item $N$: The \textit{name} of the model
    \item $C$: The \textit{context} is a set of elements of the form $(Kind, Att)$ where $Kind$ is a kind and $Att$ defines what values should have the different attributes of the instance of the kind $Kind$ before the phenomenon described by the model. The value of the attribute can be a specific constant or an unknown value.
    \item $R$: The \textit{result} is a set of elements of the form $(Kind, Att)$. They specify which attributes of which kind are changed as a result of the phenomena described by the model. For implementation purpose, we use constant to define UNSET attribute value in the \textit{context} $C$ that get set to MODIFIED values in the \textit{result} $R$
    \item $ MoF$: The \textit{model of} is to speed-up the search for models by stating which attribute of which kind is changed, without stating any specific values. It allows the system to eliminate quickly the models that are not about the right kind/attribute.
    \item $S$: The \textit{story} is the description of how we get from the \textit{context} to the \textit{result} and is in essence the core of the model. 
\end{itemize}

\subsection{Interaction pattern}
\label{sec:framework_interaction}

The dialogue with the user begins with the presentation of the output of the AI system. The user is then free to require explanations on the AI system. We authorize two types of question: the first type is requiring explanation on a value of an entity and returns Entity-Entity relations while the second is requiring explanation on a specific Entity-Entity relation and returns a model for a Model-Entity relation.

This pattern of interaction is motivated by Johnson and Johnson\cite{Johnson1993} whose survey suggests that when an expert explains a task to a novice, declarative knowledge (facts) is exchanged before procedural knowledge (how to do). We tried to match this pattern with our interaction pattern as Entity-Entity relations model in a way causal relationships while Model-Entity relations explain how a certain entity came to be from the facts.

Once the user is presented with a new Entity-Entity relation, he can query an explanation either on the relation itself, the explanan or another entity/relation that he was presented earlier in the dialogue. The user is free to continue the discussion how he likes, exploring the mental-model by himself. 

\subsection{Search for explanation}
\label{sec:framework_search}
The user being able to question any entity in the mental-model and not just the output of the system, we implemented a search for explanation to query the mental-model and return relevant information. We already highlighted that we allowed for two types of questions/explanations and both cases can be treated separately.

When asking for an explanation about a specific entity, we require the user to specify what characteristic of this entity he is interested in. We then search all the Entity-Entity relations that have this (entity, characteristic) pair as explanandum and return the ones that haven't been presented yet starting with those with the higher priority $P$. This mechanism allows for a biased presentation of Entity-Entity explanations as the ones with the highest priority are presented the first time the question is asked, and the others are only presented if the exact same question is asked again. 

When requiring an explanation about an Entity-Entity relation, the explanation system needs to search the models to find the appropriate one. As the Entity-Entity relation contains the precise information about the kind of the eplanandum and explanan they contain (with $En_t$ and $Ed_t$), we simply need to check the models : we select the one(s) where the explanandum kind is both in the context $C$ and in the result $R$ of the model and the explanan kind is in the context $C$ of the model. We then return the model to the user. We do not implement a full formulation for Model-Entity relations in the form of Model-Kind and Kind-Entity but our implementation and search algorithm both rely on these ideas.

\subsection{Examples}
\label{sec:framework_examples}

We give here a possible implementation of mental-models for two very different systems. We depict how we defined the kinds, models and entity-entity relations while the entities are instantiated from the kinds at run-time. The detailed representation of each category is given in Appendix \ref{ap:implementation}. These implementations are examples of mental-models for these systems, other choices could have been made in defining the kinds and models. 

\subsubsection{Artificial neural network}

We chose for our first example a feed-forward neural network classifier on the MNIST dataset\cite{LeCun2010}. The network is composed of three layers (numbered 0, 1 and 2) each containing 784, 30 and 10 neurons. 

\paragraph{Kinds:} We define 3 different kinds : "neuron", "parameter" and "network output".
The neuron class get instantiated once for each neuron of the network, each instance gets attributed its layer number, position in the layer together with its activation value. For each parameter in the network we instantiate the parameter kind, setting its layer and the positions of the two neurons it is connecting in their respective layers. The network output represents the final result given by the network, which is not the activation of the neurons in the output layer but the position of the biggest activation.

\paragraph{Entity-Entity relation:} We define 3 types of Entity-Entity relations : "neuron to neuron activation", "parameter to neuron activation" and "output layer to output answer". The first relation is instantiated for each link between two neurons in the neural network and represents the fact that the lower layer neuron's activation value influenced the upper layer neuron's activation value. The second relation is instantiated between each parameter and the neuron whose activation was computed using it. These relations represent the fact that the parameter influenced the activation value of the neuron. The third type of relation is instantiated between each neuron in the output layer and the network output, representing that each of these neurons were taken into account to decide the output of the classifier.

\paragraph{Models:} We define two different kinds of models: "neuron activation" and "output generation". The first type of model corresponds to the equation that computes each of the neuron's activation value from the previous layer's activation and the parameters. We then have as many models as there is of neurons in the second and third layer. The second type of models defines how the output of the network is computed from the output layer (argmax). 

\subsubsection{Prolog}

Our second example aims to implement explanations for a restriction of Prolog. It is interesting to highlight here that in contrast to the neural network example, the depiction of kinds and entity-entity relations do not represent exactly how the algorithm computed the output. This illustrates that one doesn't need to stick exactly to the AI system's computation when creating a mental-model. We parse the trace provided by the Prolog solver to create a derivation tree that we use to instantiate our explanation system. To keep this example simple, we assumed the Prolog program to be grounded, used as data a succeeding derivation tree without backtracking and no negation by failure. We note that with this structure, the entity-entity relations are going to depend on the trace and so depend on the specific Prolog query by the user. 

\paragraph{Kinds:} We define two kinds: "predicates" and "rules". A predicate is instantiated for each predicate in the program, storing a Boolean if it evaluated to true and whether it is a fact or not. A rule is also instantiated for each rule in the program and stores a Boolean indicating if the rule was used, meaning that it participated in the trace. 

\paragraph{Entity-Entity relation:} We define three different types of Entity-Entity relations: "head of rule to predicate", "predicate to body of a rule" and "fact to fact". The first relation is instantiated between a rule and a predicate, indicating that the predicate evaluated to true because it unified with the head of this used rule. The second relation is instantiated between a predicate and a rule, indicating that this specific predicate in the body participated to the body evaluating to True by itself evaluating to True. The third relation is instantiated between a predicate which is a fact and itself, indicating that it evaluated to True because it is a fact in the program. 

\paragraph{Model:} We define three different models (one for each type of relation) : "used rule", "body is true", "fact is a fact". The first model explains that when a rule is used (its body evaluated to True), the predicate in its head evaluates to True. The second model indicates that when the body of a rule evaluates to true, because all of the predicates in it evaluate to true, the rule is considered as used. The last model indicates that a predicate that is a fact in the program always evaluates to True. In this implementation, the models are exactly the abstract versions of the Entity-Entity relations.

\section{Discussion}
\label{sec:discussion}

This work proposes a general structure to implement mental-models of the AI systems. Considering the different AI architectures as entities ruled by models allows for a uniform approach to building these mental-models, instead of having to consider each specific algorithm class as a particular case requiring a unique interaction pattern. One single explanation search algorithm can be used for all the different AI systems as long as the "causal" Entity-Entity relations and models are implemented. Choosing a decomposition as relations and models represent the main difficulty when using this framework. We also define an empirical definition for explanations as the tuple composed of the explanan, explanandum, form and relation though such a classification is of little practical use outside the framework.

%\subsection{Allows for "explanation of explanation"}
% Such a architectures allows for reasoning on the explanation themselves, and questioning on the explanans without difficulty, as we now know where to look for a new cause (in the category, or with the next level of abstraction)

The explanations that our explanation system provides are Model-Entity and Entity-Entity relations which define clearly what the explanandum and explanan are. These are either an entity or a model, allowing the user to continue questioning the presented explanan if he requires more information to be satisfied: we allow for "explanations of explanation". Explanations for models are not yet implemented but they would involve other models (sub-model relation) or theories. This framework allows for a natural handling of a dialogue with the user until he is satisfied with the explanations.

%\subsection{Reusability of a modular framework}
% As we now have a modular framework, it allows for the reuse of some parts for another application. An explanation mechanism for back-propagation theory can be reused for all neural net, a statistical feature extraction from the training data explanation module can be reused regardless of the system (ANN, linear...)

%We already pointed it out in the previous section, but the classification into categories deserves to be highlighted for the fact that it renders the explanation creation modular. It defines the differences between the types of explanations we generate, and this between the algorithms that generate them. We will be able to compare algorithms belonging to the same type of explanation and possibly reuse them for other application. We allows us to classify the explanation algorithms and ease the incorporation of other type of explanation to complete our explanation system.

%\subsection{Possibility to include explanation techniques in the mental model}

Implementing a complete mental-model of the functioning of the AI system would help the user to learn the complete algorithm used for prediction but doesn't by itself assure that the user will trust the AI-system. For example knowing the forward pass algorithm of a convolutional neural network (CNN) doesn't mean that we "understand" it's output, either by lack of interpretability or other criteria. 
However, the framework allows for the inclusion of state-of-the-art explanation techniques for ML systems, suffice to find the representation of its algorithm as entities, kinds and models. For example for a CNN, one could supplement the mental-model with the information required for Layer-Wise Relevance Propagation\cite{Bach2015} (LRP) and create an Entity-Entity relation between the output of the network and the generated saliency map. This would allow the user to get an interpretation over the input while being able to inquire about the prediction and LRP algorithms. 

The classification that we operate of explanations as being of multiple forms (primarily Entity-Entity and Model-Entity in AI) is also another approach to classify AI explanations. Such a classification is not so useful for the AI community itself where we find much more appropriate classification of explanation methods\cite{Gilpin2018} for technical considerations. However for non-technical consideration, as to how to deliver explanation to non-AI experts, we hope that this work shows a different approach as to explanation definition based on social sciences. Most explanation techniques in AI being Entity-Entity relations, we believe this classification can also give hints as to where we need to focus our efforts to generate more human-like explanations.

%\subsection{Completeness and interpretability of explanations}
%The latest point made allows to discuss the notions of interpretability and completenss of explanations as defined by Gilpin et al.\cite{Gilpin2018}. 

\section{Conclusion}
\label{sec:conclusion}

In this paper we explored how one might create a system capable of generating composed explanations, which are explanations composed of multiple individual explanations obtained through a dialogue, to explain the prediction of an AI system to a non-expert user. We realized that in order to do so, we needed to implement a mental-model of the AI system that represents all the knowledge that we wish the user to have access to. We took inspiration from the structure of scientific explanation to come up with a new way of considering AI systems as an observed phenomenon composed of multiple entities interacting with one another. Such a depiction of AI algorithms allows for a general way of implementing mental-models which is agnostic to the algorithm type and allows for more natural interaction between the AI system and the user. We implemented these ideas with two examples: an artificial neural network and a simple prolog program. 

Future work can follow many tracks so we will present only a few of them. The first interesting follow-up would be to evaluate our approach through user studies. In our motivations, we made the assumption that a lay user would be in a way sensitive to scientific explanations so using its structure would make the user feel more at ease. This assumption needs to be backed up by empirical evidence to allow for a full justification of this work. However, how to define and proceed with a rigorous user study in this domain is a research line itself. 

For the system to be more accessible, it needs to be made more user-friendly with a fully developed user-interface. Our current implementation uses text in a terminal, which is enough to interact with the user and present all the needed information but hurts the fact that we would like this system to be usable by all. 

Considering the framework itself, there are multiple possible improvements. Firstly, as all Entity-Entity relations are based on a model, we could automatically generate these relations by going through the models when we initialize the entities, instead of defining each Entity-Entity relation separately. Then completing the representation of models could improve the interaction with the user. Indeed our implementation only allows for a limited amount of explanation, when a system using the full power of the structure of scientific explanation should be capable of linking models together and present theories when necessary. The same could be done for kinds, where we do not allow the user to request information about kinds and explore the characteristic of the abstract concepts. Finally, work should be done on biasing the search for explanations. Our implementation only allows for a brute force encoding of priority between explanations, when in reality we would like the bias to take into account exterior factors to display the explanation which satisfies the user more. 

\bibliographystyle{plain}
\bibliography{main}

\clearpage
\appendix

\section{Implementation details}
\label{ap:implementation}

We detail here the implementation for each kind, relation and model for the two examples in this paper. We remind from Section \ref{sec:framework} that we use the following representations : 
\begin{itemize}
    \item Kind : $(N, C, V)$
    \item Entity-Entity relation : $(N, En_t, Ed_t, En, Ed, R, P)$
    \item Model : $(N, C, R, MoF, S)$
\end{itemize}
\subsection{Neural network}

\paragraph{kinds : }

\begin{itemize}
    \item Neurons : 
        \begin{itemize}
            \item $N$ : "Neuron"
            \item $C$ : \{\}
            \item $V$ : \{"activation":FLOAT, "layer":INT, "position":INT, "name":STRING\}
        \end{itemize}
    \item Parameters : 
        \begin{itemize}
            \item $N$ : "Parameter"
            \item $C$ : \{\}
            \item $V$ : \{"value":FLOAT, "layer":INT, "i":INT, "j":INT, "name":STRING\}
        \end{itemize}
    \item Output of the network : 
        \begin{itemize}
            \item $N$ : "OutputAnswer"
            \item $C$ : \{\}
            \item $V$ : \{"value":INT, "name":STRING\}
        \end{itemize}
\end{itemize}

\paragraph{Entity-Entity relations :}
\begin{itemize}
    \item Neuron to neuron activation : 
        \begin{itemize}
            \item $N$ : "NeuronToNeuronActivation"
            \item $En_t$ : ("Neuron", activation)
            \item $Ed_t$ : ("Neuron", activation)
            \item $R$ : "This lower layer neuron's activation participated in the computation of the questioned activation"
            \item $P$ : 0
        \end{itemize}
    \item Parameter to neuron activation : 
        \begin{itemize}
            \item $N$ : "ParameterToNeuronActivation"
            \item $En_t$ : ("Parameter", value)
            \item $Ed_t$ : ("Neuron", activation)
            \item $R$ : "This parameter value participated in the computation of the questioned activation"
            \item $P$ : 0
        \end{itemize}
    \item Output Layer to Output Answer : 
        \begin{itemize}
            \item $N$ : "OutputNeuronToOutputNetwork"
            \item $En_t$ : ("Neuron", activation)
            \item $Ed_t$ : ("OutputAnswer", value)
            \item $R$ : "This lower layer neuron's activation participated in the computation of the questioned activation"
            \item $P$ : 0
        \end{itemize}
\end{itemize}

\paragraph{Models :}
We will use $l$ to represent the layer number, $i$ to represent the position of a neuron in the lower layer $l$ and $j$ to represent the position of a neuron in a upper layer $l+1$. We remind here that the output layer of the network is numbered 2. 

\begin{itemize}
    \item Neuron Activation : 
        \begin{itemize}
            \item $N$ : "Neuron activation"
            \item $C$ : \{("Neuron", ("layer"=$l+1$, "position"=j)), ("Neuron", ("layer"=$l+1$, "position"=i)), ("Parameter", ("layer"=$l$, "i"=i, "j"=j) \} for all $i$
            \item $R$ :  \{("Neuron", ("layer"=$l+1$, "position"=j, "activation"=MODIFIED)) \}
            \item $MoF$ : ("Neuron", activation)
            \item $S$ : $x_j^{l+1} = g(\sum{i} x_i^{l}+b_j^l)$
        \end{itemize}
    \item Output generation : 
        \begin{itemize}
            \item $N$ : "Output generation"
            \item $C$ : \{("Neuron", ("layer"=2, "position"=i)), ("Neuron", ("OutputAnswer", ("value"=MODIFIED))\} for all $i$
            \item $R$ :  \{("OutputAnswer", ("value"=MODIFIED)) \}
            \item $MoF$ : ("OutputAnswer", value)
            \item $S$ : $output = argmax(\{x_i^{2}/\forall i \})$
        \end{itemize}
\end{itemize}

\subsection{Prolog}

\paragraph{Kinds :}

\begin{itemize}
    \item Predicate : 
        \begin{itemize}
            \item $N$ : "Predicate"
            \item $C$ : \{\}
            \item $V$ : \{"fact":BOOL, "truth":BOOL, "text":STRING, "name":STRING\}
        \end{itemize}
    \item Rule : 
        \begin{itemize}
            \item $N$ : "Rule"
            \item $C$ : \{\}
            \item $V$ : \{"used":BOOL, "head":STRING, "body":STRING, "name":STRING\}
        \end{itemize}
\end{itemize}

\paragraph{Entity-Entity relation :}
\begin{itemize}
    \item Head of rule to predicate : 
        \begin{itemize}
            \item $N$ : "HeadToPredicate"
            \item $En_t$ : ("Rule", (used, head))
            \item $Ed_t$ : ("Predicate", truth)
            \item $R$ : "This predicate is true because it is the head of this used rule"
            \item $P$ : 0
        \end{itemize}
    \item Predicate to body of a rule : 
        \begin{itemize}
            \item $N$ : "PredicateToBody"
            \item $En_t$ : ("Predicate", truth)
            \item $Ed_t$ : ("Rule", used)
            \item $R$ : "This rule was used because this predicate in the body was true"
            \item $P$ : 0
        \end{itemize}
    \item Fact to Fact : 
        \begin{itemize}
            \item $N$ : "FactToFact"
            \item $En_t$ : ("Predicate", fact)
            \item $Ed_t$ : ("Predicate", truth)
            \item $R$ : "This predicate is True because it is a fact"
            \item $P$ : 0
        \end{itemize}
\end{itemize}

\paragraph{Model :}

\begin{itemize}
    \item Used rule : 
        \begin{itemize}
            \item $N$ : "UsedRule"
            \item $C$ : \{("Predicate", ("truth")), ("Rule", ("used"=True, "head"))\}
            \item $R$ :  \{(("Predicate", ("truth" = True))\}
            \item $MoF$ : ("Predicate", truth)
            \item $S$ : "A used rule makes the predicate in its head True"
        \end{itemize}
    \item Body is True : 
        \begin{itemize}
            \item $N$ : "TrueBody"
            \item $C$ : \{("Predicate", ("truth"=True)), ("Rule", ("Used"))\}
            \item $R$ :  \{("Rule", ("Used"=True)) \}
            \item $MoF$ : ("Rule", Used)
            \item $S$ : "A rule is considered used when each element in body evaluated to True"
        \end{itemize}
    \item Fact is a fact : 
        \begin{itemize}
            \item $N$ : "Fact"
            \item $C$ : \{("Predicate", ("truth", "fact"=True))\}
            \item $R$ :  \{("Predicate", ("truth"=True, "fact"=True)) \}
            \item $MoF$ : ("Predicate", Used)
            \item $S$ : "A predicate which is a fact in the program will always evaluate to True"
        \end{itemize}
\end{itemize}

\end{document}